\def\year{2020}\relax
\documentclass[letterpaper]{article} % DO NOT CHANGE THIS
\usepackage{aaai20}  % DO NOT CHANGE THIS
\usepackage{times}  % DO NOT CHANGE THIS
\usepackage{helvet} % DO NOT CHANGE THIS
\usepackage{courier}  % DO NOT CHANGE THIS
\usepackage[hyphens]{url}  % DO NOT CHANGE THIS
\usepackage{graphicx} % DO NOT CHANGE THIS
\usepackage{graphicx}  % DO NOT CHANGE THIS
\usepackage{latexsym}
\usepackage{enumitem}
\usepackage{url}
\usepackage{amsmath}
\usepackage{csquotes}
\usepackage{color,soul}
\title{Just Add Functions: A Neural-Symbolic Language Model}
\author{David Demeter\\
{Northwestern University}\\
{Evanston, IL, USA}\\
ddemeter@u.northwestern.edu
\And
Doug Downey\\
{Allen Institute for Artificial Intelligence}\\
{Seattle, WA, USA}\\
dougd@allenai.org} 

\begin{document}

\maketitle

\begin{abstract}

Neural network language models (NNLMs) have achieved ever-improving accuracy due to more sophisticated architectures and increasing amounts of training data.  However, the inductive bias of these models (formed by the distributional hypothesis of language), while ideally suited to modeling most running text, results in key limitations for today's models.  In particular, the models often struggle to learn certain spatial, temporal, or quantitative relationships, which are commonplace in text and are second-nature for human readers.  Yet, in many cases, these relationships can be encoded with simple mathematical or logical expressions.  How can we augment today's neural models with such encodings?

In this paper, we propose a general methodology to enhance the inductive bias of NNLMs by incorporating simple functions into a neural architecture to form a hierarchical neural-symbolic language model (NSLM).  These functions explicitly encode symbolic deterministic relationships to form probability distributions over words.  We explore the effectiveness of this approach on numbers and geographic locations, and show that NSLMs significantly reduce perplexity in small-corpus language modeling, and that the performance improvement persists for rare tokens even on much larger corpora.  The approach is simple and general, and we discuss how it can be applied to other word classes beyond numbers and geography. 

\end{abstract}

\section{Introduction}
Neural network language models (NNLMs) have achieved ever-improving accuracy due to more sophisticated architectures and increasing amounts of training data \cite{radford2019language,Krause2019DynamicEO,Merity2017RegularizingAO}.  These models are formulated on the inductive bias of the distributional hypothesis of language, which states that words appearing in similar contexts are likely to have similar meanings \cite{firth57synopsis}.  This is realized in NNLMs by embedding words with similar meanings close to each other in a high-dimensional space.

\begin{figure}[!ht]
\includegraphics[scale=0.65]{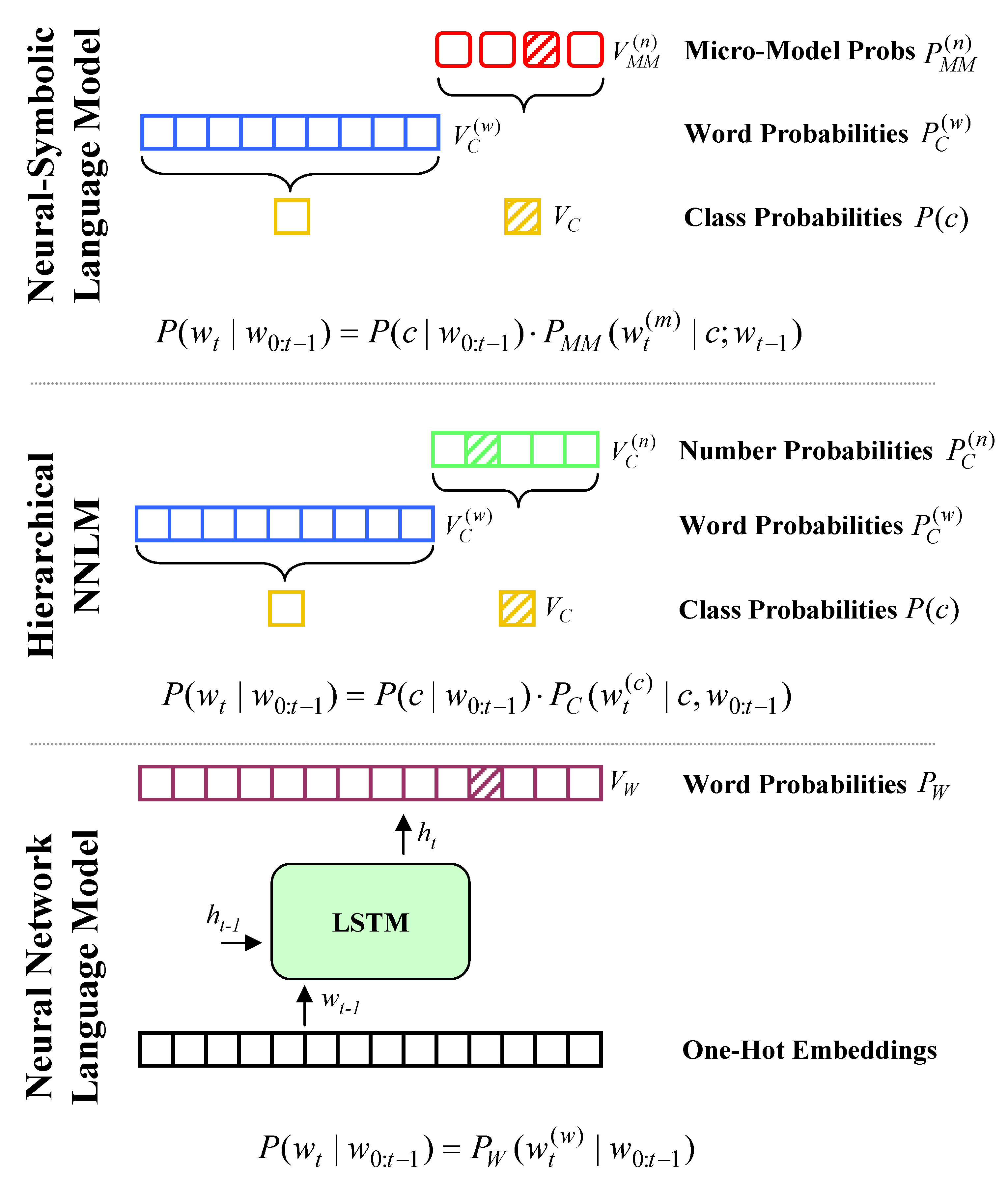}
\centering
\caption{\textbf{Neural-Symbolic Language Model Architecture.} %{The bottom panel shows a traditional NNLM where a LSTM generates a probability distribution $P_W$ over the entire vocabulary $V_W$.  The middle panel shows a hierarchical NNLM where the probability of token $w_t$ is the product of the word class probability $P(c)$ and the probability distribution $P_C$ over the word class vocabularies $V_C^{(w)}$ and $V_C^{(n)}$.  The top panel shows the architecture of a NSLM, where the neural method of generating $P_C$ over $V_C^{(n)}$ is replaced with a micro-model using simple functions to generate $P_{MM}$, and the probability of token $w_t$ is the product of the word class probability $P(c)$ and the probability distribution $P_{MM}$ over the class vocabulary $V_{MM}$.}
{The bottom panel shows a traditional NNLM where a LSTM language model generates a probability distribution $P_W$ over the entire vocabulary $V_W$ (the purple cells).  The middle panel shows a hierarchical NNLM where a probability distribution $P(c)$ is generated over a word class vocabulary $V_C$ (the yellow cells) and probability distributions $P_C^{(w)}$ and $P_C^{(n)}$ over disjoint word and number vocabularies $V_C^{(w)}$ (the blue cells) and $V_C^{(n)}$ (the green cells), respectively. Up to this point, all probabilities are estimated using neural methods.  The top panel depicts a NSLM where probability $P_C^{(n)}$ is replaced with a micro-model (the red cells) which uses symbolic expressions to generate a probability distribution $P_{MM}^{(n)}$ over vocabulary $V_{MM}^{(n)}$ of number tokens.    In the {\em William the Conqueror} example, all tokens are members of $V_W$, non-number tokens are members of  $V_C^{(w)}$, number tokens are members of $V_C^{(n)}$, while 1066 and 1087 map to $w_{t-1}^{(n)}$ and $w_{t}^{(n)}$, respectively to form $V_{MM}^{(n)}$.}
}
\label{fig:arch}
\end{figure}

The NNLM inductive bias, while well-suited to modeling most running text, can result in key limitations for the models.  In particular, today's NNLMs often struggle to learn certain spatial, temporal, or quantitative relationships, that are commonplace in text and are second-nature for human readers.  Consider the following examples in which a language model is asked to predict the final word shown in square brackets:

\begin{itemize}
\setlength\itemsep{0.20em}
\item [] \textit{William the Conqueror won the Battle of Hastings in 1066, and died in [1087]} 
\item [] \textit{Exciting European cities are Paris, Rome and [London]}
\end{itemize}

Based on the context, a human reader can predict that the target word in the first case will likely be a four-digit number that is at least as large as (but not much larger than) 1066. Likewise, the target word in the second case will be a city that is geographically close to the other locations (i.e., a city in Europe). While these expectations can be encoded with simple mathematical or logical expressions, as we show in our experiments they are nontrivial for NNLMs to learn.

How can we augment today's neural models with such encodings?  In this paper we propose a general methodology to enhance the inductive bias of NNLMs by incorporating simple functions into a neural architecture to form a hierarchical {\em Neural-Symbolic Language Model} (NSLM).  To formulate a NSLM, words associated with a particular language phenomena are aggregated into a small number of classes with their own class-specific vocabularies (e.g., four-digit years).  The NSLM takes the form of a hierarchical NNLM \cite{Bengio2006NeuralPL} that jointly generates a probability distribution over the classes and the non-class vocabulary.  Simple functions are then used to allocate probability assigned to a particular class to the words in the class-specific vocabulary (see Figure \ref{fig:arch}).  These simple functions explicitly encode symbolic deterministic relationships to form probability distributions over a class-specific vocabulary, and are called {\em micro-models} (MMs).  For example, an MM for four-digit years could capture how the numeric difference between two consecutive years in text often follows a predictable distribution (see Figure \ref{fig:gauss}). 

We implement our approach within Recurrent NNLMs, and demonstrate the effectiveness of NSLMs in the number and geographic domains. Micro-models are formulated using a small set of general-purpose building blocks and parameters of these functions are learned on the training corpora. We show that NSLMs reduce perplexity for numbers and geographic locations (4\% of the tokens in our corpora) by approximately 40\% and 60\%, respectively, on the Wikitext-2 data set, and outperform the previous state-of-the-art for modeling numbers within a language model \cite{Spithourakis2018NumeracyFL}. While NSLMs are most useful for small training corpora, we show that they can still offer significant advantages on the larger Wikitext-103 data set, especially for rare tokens.  The approach is simple and general, and we discuss how it can be applied to other word classes.

\section{Previous Work}

Our work is related to three active subareas within language modeling research: numeracy, predicting geographic locations and integrating symbolic reasoning with NNLMs.

Numeracy in language models has only recently started to be explored.  Spithourakis \textit{et al.} \shortcite{Spithourakis2016NumericallyGL} showed how to ground numbers in language models by appending the value of a number to its word embedding.  A variety of techniques to improve numeracy in language models was more recently explored, including hierarchical models, character models, mixture of Gaussians and hybrid models of these techniques \cite{Spithourakis2018NumeracyFL}.  These models were conditioned using only the preceding textual context and emphasized the generation of out-of-vocabulary numbers.  Other related work explores the use of language models to solve algebra word problems \cite{Kushman2014LearningTA,Roy2015SolvingGA} or to evaluate nested arithmetic expressions \cite{Hupkes2018VisualisationA}.  By contrast, we model numbers in general-domain text, and use this as a test bed for new methods that incorporate symbolic functions into neural models.

Most works considering geographic locations in language models are focused on geotagging social media content.  {\em Geotagging} is generally defined as estimating the geographic coordinates of a piece of content such as text, an image, etc., \cite{KordopatisZilos2017GeotaggingTC}.  Geotagging approaches often rely on gazetteers to provide textual features \cite{Smart2010MultisourceTD} and large-scale language modeling approaches \cite{OHare2012ModelingLW}.

The closest prior work to our own are methods that incorporate symbolic knowledge into NNLMs.  One of the more prominent examples is the Neural Knowledge Language Model (NKLM), which composes a knowledge graph with a NNLM \cite{Ahn2017ANK}.  This knowledge is acquired symbolically from a plain text triple (subject, relation, object) and is encoded in an embedding space.  The model is trained to predict \enquote{fact} or \enquote{not a fact}, in which case the next token in a sequence is selected from the knowledge triples or the word vocabulary.  This work was premised on improving performance on rare tokens, and in particular named entities.  Neural-symbolic integration approaches \cite{Besold2017NeuralSymbolicLA} explore general methods of integrating first-order logic with neural networks.  Our work is less broad and focuses on formulating NSLMs that can encode inductive bias for specified domains.

Our approach is distinct from each of these in that we encode an enhanced inductive bias by appending functionality to a NNLM in the form of symbolic functions, which are probabilistically triggered in response to the neural outputs.

\section{Models}

We now formally introduce the components of an NSLM and discuss how class-specific NSLMs are formulated.

\subsection{Background}

A {\em Neural-Symbolic Language Model} (NSLM) is a hierarchical NNLM that incorporates simple functions to enhance inductive bias.  Formally, the NSLM is defined over one or more classes of tokens $C$, which each represent natural language phenomena incompletely captured by a traditional NNLM.  Individual tokens $w_t^{(c)}$ comprising each class form a vocabulary $V_C$ unique to $C$.  A NSLM consists of two components: a hierarchical NNLM which assigns probability to each class $C$, and a micro-model that allocates this probability over the words in $V_C$.  The hierarchical NNLM operates over individual word tokens and special $<$tag$>$ tokens used to label each class $C$.  A {\em micro-model} is a simple symbolic function that encodes an inductive bias specialized to a given class $C$, aimed at improving the model's predictive performance on tokens from $C$.

In general, micro-models can be manually specified using domain knowledge about the classes $C$.  In this work, we find that micro-models for our target domains can be built up from a small set of simple standardized components.  These components fall into two categories: metric functions that calculate a scalar $m_t^{(c)}$ for each possible token $w_t^{(c)}$ in position $t$ within the class vocabulary $V_C$, and a probability density function (PDF) that maps all $m^{(c)}$ to a probability distribution over $V_C$.
The metric functions that we work with are either unary $f_c(w_t^{(c)})$ and consider only the token at position $t$, or binary $f_c(w_t^{(c)},w_{t-1}^{(c)})$  and operate over both the current token $w_t^{(c)}$ and the immediately preceding token $w_{t-1}^{(c)}$ belonging to $C$.

\begin{figure}[ht]
\includegraphics[width=3.0 in]{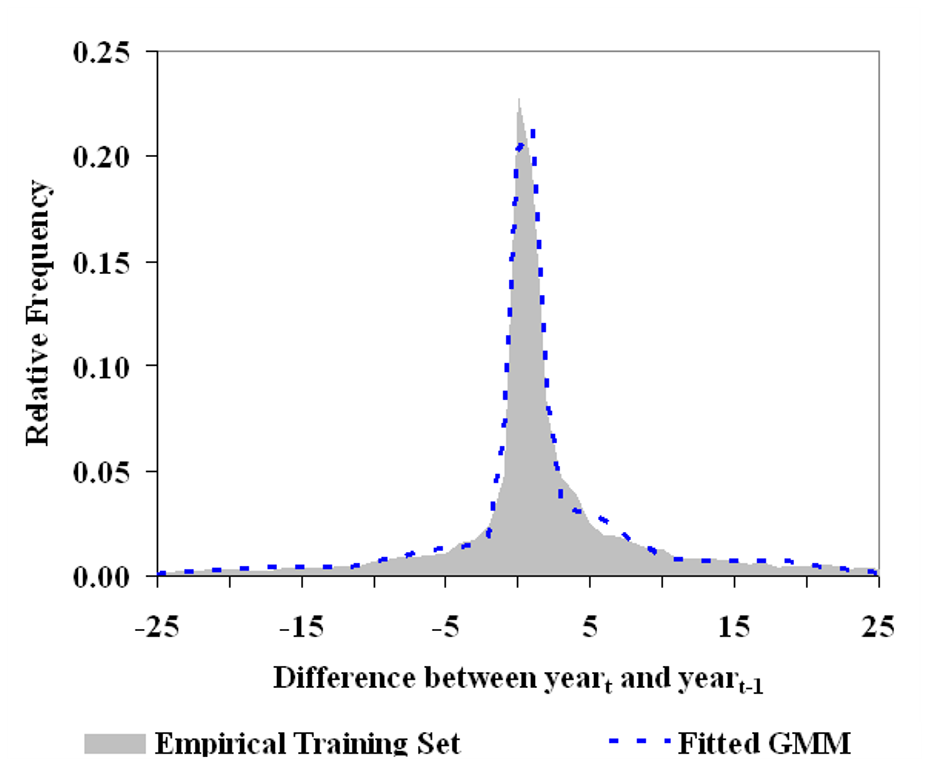}
\caption{\textbf{Fitted Gaussian for $<$year$>$ Class.}  {The distribution of the difference between the token $year_t$ and the preceding year token $year_{t-1}$ is highly peaked at one year, illustrating the strong locality of this number class.}}
\label{fig:gauss}
\end{figure}

Number tokens in natural language text offer a concrete example of where NSLMs could be helpful.  Traditional NNLM perform poorly over number tokens \cite{Spithourakis2018NumeracyFL} in that they do not fully capture the phenomenon that numbers can be expressed as ordered sets.  Specifically, consider $<$year$>$ tokens in natural language text forming a class $C$.  The vocabulary $V_C$ would consist of all four-digit years in a training corpus.  In the corpora that we worked with, we observed strong locality among year tokens, meaning that the four-digit year most likely to occur next in text is based on the numeric interval between it and the year occurring most recently in the context (see Figure \ref{fig:gauss}).  This regularity, while difficult for an NNLM to learn, forms an inductive bias that can be encoded with a micro-model where the metric function calculates the difference between two consecutively occurring years and the PDF maps these differences to a probability distribution over all four-digit years in the vocabulary $V_C$. 

\subsection{Micro-Models}

We define a micro-model as a compact expression that uses simple functions to generate a probability distribution $P_{MM}$ over a vocabulary $V_{MM}$ to predict word $w_{t}^{(m)}$, such that:

\begin{equation} \label{eq:mm}
P_{MM}(w_t^{(m)}|c;w_{t-1}^{(m)})=MM(w_{t}^{(m)},w_{t-1}^{(m)},\theta_{MM})
\end{equation}

\noindent where $c$ is the word class, $MM$ is a symbolic expression which implements and applies $f_c$ to a PDF to generate a probability distribution, $\theta_{MM}$ is a set of parameters learned from data, and $w_{t-1}^{(m)}$ and $w_{t}^{(m)}$ are members of the same vocabulary $V_{MM}$ and $w_{t-1}^{(m)}$ is the most recent occurrence of a class $m$ token preceding $w_{t}^{(m)}$ (e.g., in the illustration above, $1066$ and $1087$ would be examples of $w_{t-1}^{(m)}$ and $w_{t}^{(m)}$, respectively).

\subsection{Language Models}
RNNs with LSTM cells are a canonical architecture that, until recently, powered state-of-the-art performance on most corpora \cite{Merity2017RegularizingAO}.  Since the micro-model approach is equally applicable to any other language modeling architecture that uses a softmax output layer, we use LSTMs for evaluation.  An LSTM assigns a probability distribution $P_W$ over a vocabulary $V_W$ to predict the next word $w_t$ in a sequence by jointly learning neural network parameters $\theta_{NNLM}$ and word embeddings $\boldsymbol{W}$ to produce a hidden state $h_t$ that summarizes the previous context $w_{0:t-1}$.  

\begin{align}\label{eq:lstm}
P(w_t | w_{0:t-1}) = P_W(w_t | h_t ; {\bf \theta_{NNLM}})
\end{align}

In hierarchical NNLMs \cite{Morin2005HierarchicalPN}, tokens are assigned to two or more word classes $C$ consisting of class-specific vocabularies $V_C$, which are typically disjoint subsets of $V_W$.  The NNLM architecture is modified to separately generate probability distributions $P(c)$ conditioned on $w_{0:t-1}$, and $P_C$ over the word class vocabularies $V_C$ using the neural approach described above.  The probability of an individual token is the product of these two distributions:

\begin{multline}\label{eq:h-nnlm}
P(w_t|w_{0:t-1}) = P(c|w_{0:t-1}) \cdot P_C(w_t^{(c)}|c;w_{0:t-1})
\end{multline}

\noindent where $w_t^{(c)}$ denotes a token at position $t$ from the vocabulary $V_C$.

Our proposed architecture replaces the neural component used to generate $P_C$ with a probability distribution $P_{MM}$ derived using a micro-model, such that:

\begin{multline}\label{eq:nnlm+mm1}
P(w_t|w_{0:t-1}) = P(c|w_{0:t-1}) \cdot P_{MM}(w_t^{(m)}|c;w_{t-1}^{(m)})
\end{multline}

\noindent creating a NSLM.  Most hierarchical language models are motivated by the desire to reduce model computational complexity.  This is often at the expense of accuracy since it requires the product of probabilities from two distributions.  By contrast, NSLMs are aimed at better capturing particular domain-specific phenomena in language.  They have a symbolic component which is conditionally fired based upon the prediction of a class within the hierarchy.  This symbolic component makes them distinct from HNLMs and allows them to improve accuracy rather than just reducing complexity. 

We believe that NSLMs provides several advantages over traditional NNLM architectures.  As simple functions, micro-models are vastly more parameter efficient than neural approaches, where even small NNLMs can have large parameter spaces.  Whereas other approaches require that external knowledge be converted into a neural form such as embeddings \cite{Ahn2017ANK}, micro-models can work with knowledge in symbolic form.  In addition, micro-models are fully inspectable, addressing opacity as one of the prominent criticisms of neural approaches \cite{Marcus2018DeepLA}.

\subsection{Number and Geography NSLMs}

We evaluate NSLMs on number and geographic location word classes.  Micro-models are formulated using a small set of metric functions and PDFs.  A \textit{difference} metric function is used to calculate the numeric difference between the value of two tokens.  Unary metric functions \textit{frequency} and \textit{value} compute the frequency and numeric value, respectively, for individual tokens.  The \textit{convert} metric function calculates an exact value base upon tokens $w_{t-1}^{(c)}$ and its accompanying units to produce a \textit{correct/incorrect} metric which is evaluated against a binary distribution.  The \textit{Euclidean} metric function calculates the negative squared distance between pairs of locations using longitude and latitude coordinates.  This metric leverages Tobler's First Law of Geography \cite{Tobler1970ACM} which states that "everything is related to everything else, but near things are more related than distant things."   The quantities produced by these metric functions are mapped to probability distributions using multinomial, unigram, Gaussian and a mixture of Gaussians PDFs, which may be continuous or discrete.

An exhaustive grid search over \{metric function, PDF\} pairs is performed on the validation set to select specific micro-models.  Metric function and PDF selections for each  word class are presented in Tables \ref{tab:numresults} and \ref{tab:georesults} below.

\subsection{Generalization of NSLMs}

Despite the potential complexity of the foregoing, NSLMs are relatively straight-forward to construct (see Table \ref{tab:procedure}).  Specifying the set of candidate metric functions provides the most flexibility for authors of a NSLM.  The candidate metric functions used for our tasks are by no means exhaustive, and additional metric function can readily be specified.  Longitude and latitude values are provided to the \textit{Euclidean} metric function for geographic word classes.  Similar quantitative values can be specified (or learned) for members of other word classes.  In addition, the list of PDFs  could readily be expanded beyond those that we considered.

\begin{table}[h]
\centering
\begin{tabular}{rl}
\hline
\rule{0pt}{3ex} & {\bf NSLM Construction Algorithm} \\
\hline
\rule{0pt}{3ex}1.  & Identify a natural language phenomenon  \\
                   & associated with a class of tokens $C$, which has\\
                   & an inductive bias that can be encoded as a simple \\
                   &  mathematical or logical expression.\\
               2.  & Write a regular expression to identify tokens in \\
                   & class $C$, and let $V_C$ be the subset of the \\
                   & vocabulary that matches the regular expression. \\
               3.  & Add/delete candidate metric functions  \\
                   &  $f_c(w_t^{(c)},w_{t-1}^{(c)})$ and PDFs to be considered. \\
               4.  & Train a hierarchical NNLM on words and class\\
                   & labels and select a \{metric function,PDF\} pair.\\
\hline

\end{tabular}
\caption{\textbf{General Steps Required to Construct a NSLM.}  NSLMs can be built by applying a small number of standard language modeling tasks, and providing code to implement simple metric functions and PDFs.}
\label{tab:procedure}
\end{table}

To illustrate the creation of a NSLM for another domain, consider a language model to be trained on a recipe corpus where the objective is to improve performance when predicting verbs.  The causal ordering of verbs in recipes (e.g., ``measuring'' occurs before ``mixing,'' which occurs before ``baking,'' ``cooling,'' etc.) provides an inductive bias that can be encoded with a micro-model (Step 1).  Tokens in this word class $C$ and the class vocabulary $V_C$ could be identified with regular expressions built from a gazetteer and morphological rules (Step 2).  To encode the inductive bias as a micro-model, we could define a simple metric function that encodes whether a verb obeys the causal partial order in the cooking domain (e.g., ``cooling'' after ``baking'' obeys the partial order, but ``measuring'' after ``broiling'' does not).  A multinomial density function could then be used to assign higher probability to verbs that obey the partial order than those that do not.  We may also explore other candidate metric and PDFs  (Step 3).  Finally, the hierarchical NNLM would be trained on class labels and words, and a \{metric function, PDF\} pair would be selected to form a micro-model (Step 4).

We believe that any number of NSLMs could be similarly constructed given sufficient domain knowledge and the ability to identify an incremental inductive bias.  Other examples of potential NSLMs include (i) baseball commentary, (ii) financial reports and (iii) technical manuals.  A NSLM could be constructed to improve the running commentary of a baseball game by inducing a symbolic model of game-play from structured and unstructured data such as game states and play-by-play descriptions, respectively.  Augmenting language models with symbolic accounting and financial reporting rules could enhance performance in the growing field of text analysis in finance research.  Text generation for technical manuals could be enhanced by leveraging part-whole information about a product, function of components, etc., by retrieving these relations from a symbolic external knowledge base.

In our experiments, NNLMs were trained with standard techniques and a exhaustive grid search was used to select micro-models.

\section{Experimental Setup}

We now present our experimental setup for evaluating NSLMs on number and geographic location tokens.

\subsection{Training Corpora}

Evaluating on numbers and geographic locations requires some corpus selection and preparation.  Pre-processing applied to most popular text corpora have eliminated number tokens by replacing them with specialized tokens such as \textit{N} (Penn Treebank), word forms (text8) or tokenized representations (Wikitext).  These modifications improve overall language performance by reducing the vocabulary and removing hard-to-predict tokens.  To the best of our knowledge, the 1B Word corpus \cite{Chelba2013OneBW} is the only major text corpus which keeps numbers in their original form.  However, the 1B Word corpus is randomized at the sentence level destroying  much of the number locality.

To evaluate numbers on the Wikitext corpora, we reverse the tokenization applied to commas and decimals, and leave negative sign tokenization in place since it appears in other contexts such as hyphenated words.  Numbers expressed as words or ordinals were also ignored since their usage as numbers is ambiguous.  To evaluate geographic locations on the Wikitext corpora, multi-word named entities appearing in the Geonames data set \cite{geonames} are chunked together to form single tokens.  To minimize polysemy-related errors, only cities with populations over 500,000 are considered.  These procedures result in distinct corpora for numbers and geographic locations, with differences to the standard corpora.

\begin{table*}[ht!]
\centering
\small
\begin{tabular}{l@{\hspace{5pt}}@{\hspace{4pt}}ccrrrrrrrr}
\hline
& {\bf Metric} &  & \multicolumn{4}{c}{\bf Wikitext-2} & \multicolumn{4}{c}{\bf Wikitext-103} \\
{\bf Class} & {\bf Function} & {\bf PDF} & {\bf NNLM}& {\bf HNLM} & {\bf CRNN} & {\bf NSLM} & {\bf NNLM} & {\bf HNLM} & {\bf CRNN} & {\bf NSLM} \\

\hline
\rule{0pt}{3ex} \bf{Word Classes} \\
\hspace{0.05in} $<$year$>$ & {\em diff} & {\em MoG} & 486.3&  452.8 & 396.3 & 247.3 & 173.0 & 156.1 & 147.0 & 92.6  \\
\hspace{0.05in} $<$day$>$ & {\em freq} & {\em unigram} & 76.2 & 78.5 & 72.6 & 70.8 & 51.7 & 49.7 & 48.2 & 48.2 \\
\hspace{0.05in} $<$round$>$ & {\em value} & {\em multinomial} & 850.0 & 866.5 & 806.8 & 781.7 & 307.8 & 263.3 & 251.1 & 277.7 \\
\hspace{0.05in} $<$decimal$>$ & {\em diff} & {\em MoG} & 10,476.4 & 7,524.4 & 6,715.5 & 6,998.0   & 11,244.6 & 9,423.2 & 7,305.5 & 9,144.2 \\
\hspace{0.05in} $<$range$>$ & {\em diff} & {\em miltinomial} & 185.5 & 206.7 & 76.5 & 89.3  & 81.0 & 81.8 & 62.8 & 51.9 \\
\hspace{0.05in} $<$convert$>$ & {\em convert} & {\em binary} & 351.9 & 303.7 & 93.3 & 33.3  & 157.4 & 147.4 & 129.4 & 20.1 \\
\hspace{0.05in} $<$other$>$ & {\em value} & {\em multinomial} & 2,819.9 & 2,785.0 & 1,659.7 & 1,778.6  & 1,022.8 & 933.2 & 865.1 & 954.0 \\

\rule{0pt}{3ex} \bf{Aggregate}  \\
\hspace{0.05in} Numbers &  &  & 790.2 & 765.5 & 559.3 & 475.5 & 364.6 & 332.0 & 305.5 & 262.0 \\
\hspace{0.1in} \% change & & &  & (3.1\%)  &  (29.2\%) & (39.8\%) & & (8.9\%) & (16.2\%) & (28.1\%) \\
\hspace{0.05in} Global & & & 92.0 & 90.9 & 90.2 & 89.8 & 57.8 & 58.0 & 57.9 & 57.7 \\
\hspace{0.1in} \% change & & & & (1.2\%)  & (1.9\%) & (2.3\%) & & 0.5\% & 0.2\% & (0.2\%) \\ 
\hspace{0.05in} Num Cache & & & 508.5 & 474.9 & 374.1 & 324.4 &302.7 & 296.1 & 286.5 & 251.0 \\
\hspace{0.1in} \% change & & & (35.6\%) & (39.9\%)  & (52.7\%) & (59.0\%) & (17.0\%) & (18.8\%) & (21.4\%) & (31.2\%) \\ 

\hline
\end{tabular}
\caption{\textbf{Perplexities for Number Word Classes.} The NSLM architecture outperforms all other methods on both the Wikitext-2 and Wikitext-103 data sets, while Global perplexity does not increase indicating that the performance improvements achieved by the NSLM approach is not the result of \textit{stealing} probability mass from other parts of the model.}
\label{tab:numresults}
\end{table*}

\begin{table*}[ht]
\centering
\small
\begin{tabular}{l@{\hspace{5pt}}@{\hspace{4pt}}ccrrrrrrrr}
\hline
& {\bf Metric} &  & \multicolumn{4}{c}{\bf Wikitext-2} & \multicolumn{4}{c}{\bf Wikitext-103} \\
{\bf Class} & {\bf Function} & {\bf PDF} & {\bf NNLM}& {\bf HNLM} & {\bf CRNN} & {\bf NSLM} & {\bf NNLM} & {\bf HNLM} & {\bf CRNN} & {\bf NSLM} \\

\hline
\rule{0pt}{3ex} \bf{Word Classes} \\
\hspace{0.05in} $<$city$>$ & {\em Euclidean} & {\em Gaussian} & 24,870.5 & 19,853.3 & 26,493.1 & 2,400.4 & 2,202.7 & 1,617.4 & 1,457.6 & 1,517.1 \\
\hspace{0.05in} $<$state$>$ & {\em Euclidean} & {\em Gaussian} & 3,137.1 & 2,726.5 & 2,978.8 & 1,967.2 & 667.1 & 639.5 & 670.7 & 595.8 \\
\hspace{0.05in} $<$country$>$ & {\em Euclidean} & {\em Gaussian} &3,361.3 & 3,060.2 & 3,330.8 & 1,498.6 & 274.4 & 218.9 & 231.7 & 233.5 \\

\rule{0pt}{3ex} \bf{Aggregate}  \\
\hspace{0.05in} Locations  & & & 4,812.1 & 3,911.1 & 4,500.0 & 1,801.1 &  585.7 & 496.0 & 507.6 & 488.6 \\
\hspace{0.1in} \% change & & & & (17.1\%) & (6.5\%)  &  (62.6\%) & & (15.3\%) & (13.3\%) & (16.6\%) \\
\hspace{0.05in} Global &  & & 91.0 & 91.2 & 91.3 & 90.5 & 58.0 & 58.1 & 58.2 & 58.1 \\
\hspace{0.1in} \% change &  & & & 0.3\%  & 0.4\% & (0.5\%) & & 0.2\% & 0.3\% & 0.2\% \\ 
\hspace{0.05in} Geo Cache &  &  & 880.6 & 739.1 & 976.6 & 656.7 & 247.3 & 249.0 & 266.6 & 269.9 \\
\hspace{0.1in} \% change & & & (81.7\%) & (84.6\%)  & (79.7\%) & (86.4\%) & (57.8\%) & (57.5\%) & (54.5\%) & (53.9\%) \\ 

\hline
\end{tabular}
\caption{\textbf{Perplexities for Geographic Location Word Classes.} The NSLM architecture outperforms all other non-cache methods on both the Wikitext-2 and Wikitext-103 data sets, with Global perplexity remaining constant.}
\label{tab:georesults}
\end{table*}

\subsection{Model Parameters and Baselines}
The goal of our experiments is not to set new benchmarks for predicting number and geographic word classes in text, but instead to demonstrate that the NSLM approach can be used to improve popular neural models of language by enhancing the inductive bias.  Thus, we adopt a standard language model architecture as our primary baseline, an RNN with LSTM cells and hyper-parameters corresponding to \textit{medium} 650 dimensional models \cite{Zaremba2014RecurrentNN}.  All models converged within 20 training epochs.  During training, the softmax is computed using the full vocabulary, except for the Wikitext-103 model which uses a sampled-softmax \cite{Jean2015OnUV} with a sampling rate of 2,500.  Distributions used by the micro-models are learned on the training sets.

We evaluated against four baselines, with each successively incorporating elements of our proposed architecture:

\begin{itemize}[leftmargin=*]

\item A traditional LSTM ({\bf NNLM}) is used to assign probabilities to word classes over the vocabulary $V_W$ (Eq. \ref{eq:lstm}).
\item A hierarchical NNLM ({\bf HNLM}) is used to assign probabilities to \textit{both} word classes and members of the class vocabulary $V_C$ (Eq. \ref{eq:h-nnlm}).  This evaluates how the benefit of working with smaller vocabularies for numbers and geographic locations may offset potentially poor accuracy of class probability estimates.
\item Character RNNs ({\bf CRNN}) are used to assign probabilities to word lass tokens by predicting one character of a token at a time.  Separate CRNNs were trained for each class and probability was assigned to each token using the chain rule \cite{Jzefowicz2016ExploringTL}.  
\item {\bf Neural cache} models \cite{Grave2017ImprovingNL} consider locality in language model results by up-weighting the probability of target words for repetitions in a historical window.  Neural-cache was applied to the three other neural baselines.  \textit{Step size}, ensembling factor $\lambda_{Cache}$ and temperature $\theta_{Cache}$ were set to 500, 0.25 and 0.75, respectively, after tuning on the validation set.
\end{itemize}

\subsection{Model Implementations}

We trained separate models for number and geographic word classes.  NNLMs trained on the Wikitext-2 and Wikitext-103 corpora were on the order of 50 million and 370 million parameters, respectively.  We trained our models in a multi-task configuration, allowing us to maintain a consistent number of parameters across NNLM, HNLM and NSLM architectures.

Because our micro-models target specific phenomena, they do not form a complete and accurate distribution on their own.  Thus, the final evaluation of our proposed architecture takes the form of an ensemble between the standard NNLM (Eq. \ref{eq:lstm}) and NSLM (Eq. \ref{eq:nnlm+mm1}).  The multi-task model allows us to construct these ensembles \enquote{for free} in that the number of parameters does not increase.

\section{Results}

Our primary experimental results are shown in Tables \ref{tab:numresults} and \ref{tab:georesults}.  We report results in both a with- and without-cache setting.  Without cache, our methods beat the baseline on all four data sets, by margins ranging from 16.6\% to 62.6\%, making it the best performing of the methods we evaluated.  With cache, our approach is better on three of the four data sets, but is outperformed by NNLM for the large data set on geographic locations.  Neural methods appear capable of learning some of the functionality captured by the micro-models given additional training text, which is exhibited on the large data set by narrowing performance margins between our methods and the NNLM baselines.

Importantly, we see that using micro-models does not increase global perplexity compared with baselines across all data sets.  This shows that micro-model improvements in number and geographic location tokens do not arise simply from the model shifting probability mass away from other word classes.

%\subsection{Comparison with Prior Work}

For completeness, we also evaluated against a variety of other models proposed to improve numeracy in language models \cite{Spithourakis2018NumeracyFL}.  These models primarily explored open vocabulary methods to address the out-of-vocabulary problem associated with numeracy in language models, which is different from our task.  The NSLM architecture shows superior performance (see Table \ref{tab:riedel}).  We are not aware of any comparable baseline in prior work for geographic locations.

\begin{table}[h]
\small
\centering
\begin{tabular}{ccccc}
\begin{tabular}{lrrrr}
\hline
\rule{0pt}{3ex} {\bf }  &  {\bf Num}& {\bf \%}\\
{\bf } & {\bf PPL}& {\bf Change}\\\hline
\rule{0pt}{3ex} \bf{Spithourakis Models} \\
\rule{0pt}{3ex}\hspace{0.1in} softmax (baseline) &  709.6 &  \\
\hspace{0.1in} softmax+RNN &  700.5 & (1.3\%) \\
\hspace{0.1in} h-softmax &  761.4 & 7.3\% \\
\hspace{0.1in} h-softmax+RNN &  631.5 & (11.0\%)\\
\hspace{0.1in} MoG &  909.4 & 28.2\% \\
\hspace{0.1in} d-RNN &  1,247.8 & 75.8\%\\
\hspace{0.1in} combination &  997.0 & 40.5\% \\
\rule{0pt}{3ex} \bf{Our Model} \\
\rule{0pt}{3ex}\hspace{0.1in} NNLM (baseline) &  831.2 &  \\
\hspace{0.1in} NSLM &  494.9 & (40.6\%) \\
\hline
\end{tabular} & 
\end{tabular}
\caption{\textbf{Comparison with Other Number Class Models.} NSLMs outperform a variety of other numeracy models on Wikitext-2 with $d=300$ \cite{Spithourakis2018NumeracyFL} measured by percent improvement. Two separate baselines were presented since the numeracy models use an artificially small vocabulary, making a direct comparison difficult.}
\label{tab:riedel}
\end{table}

%\subsection{Analysis of Performance by Rarity}

\begin{figure}[h]
\includegraphics[width=2.8 in]{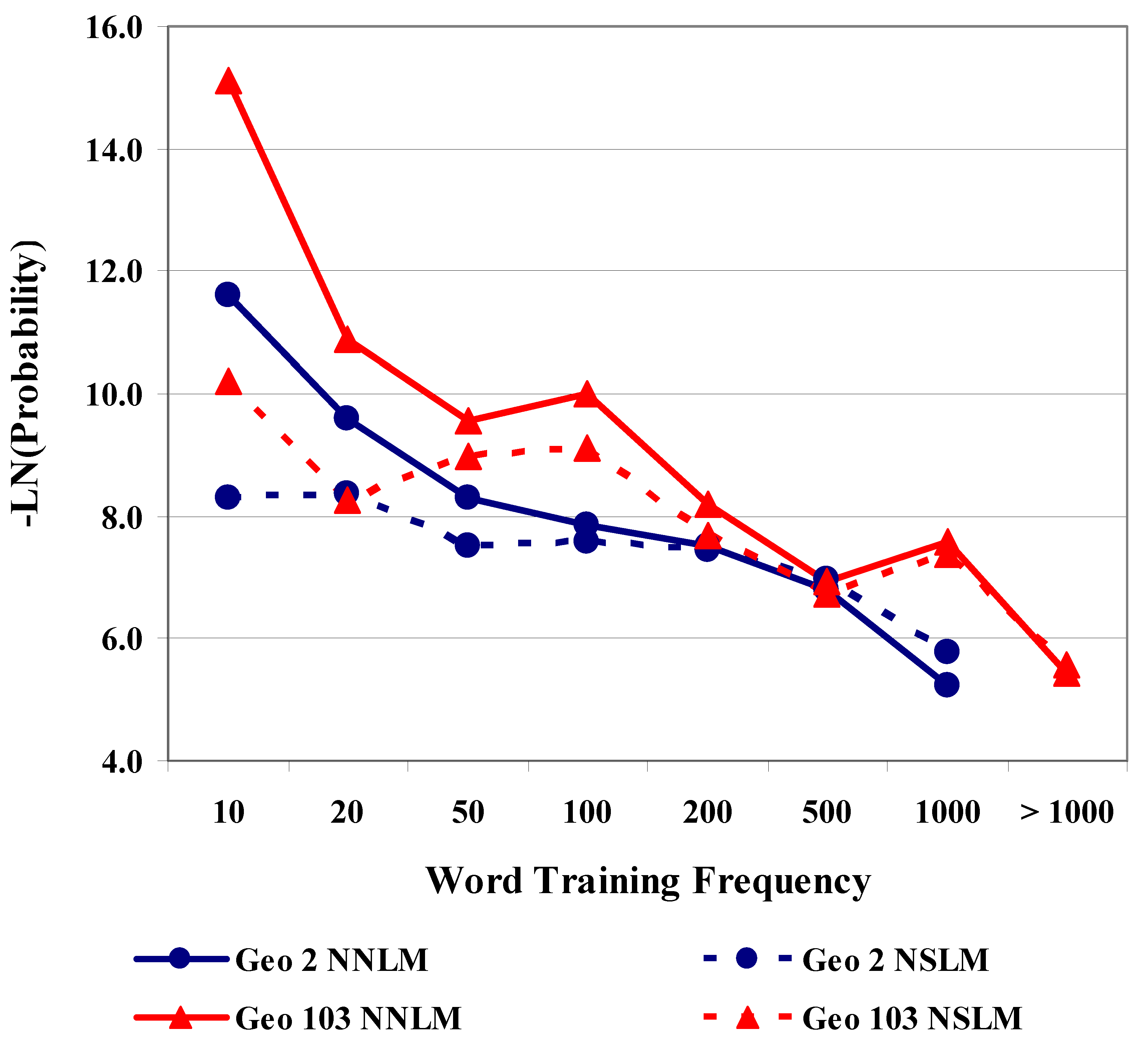}
\caption{\textbf{Performance by Rarity on Geographic Tokens.}}
\label{fig:rare2}
\end{figure}

Performance of our method on the larger data set dropped substantially for geographic locations.  However, we show that our method still substantially outperforms the NNLM baseline for rare tokens (see Figure \ref{fig:rare2}) even on the large training corpus.  This suggests that the micro-model approach can continue to offer advantages for rare tokens, even for larger corpora.

\subsection{Incorporating Large-Scale Semantics}

Our results show that NNLMs benefit from more training data, meaning that we may be evaluating against a weak baseline on the small data set.  To evaluate this potential factor, we initialized our models with GloVe embeddings trained on the 6B token corpus \cite{Pennington2014GloveGV}.  Results show a 23.0\% and 21.8\% improvement (see Table \ref{tab:glove}) on numbers and geographic locations, respectively, confirming the benefit of more training data.  However, even with stronger baselines our methods still perform better with improvements of 42.1\% and 63.1\% on numbers and geographic locations, respectively.

\begin{table}[h]
\centering
\small
\begin{tabular}{cccc}
\begin{tabular}{lrrr}
\hline
\rule{0pt}{3ex} {\bf } & {\bf NNLM} & {\bf GloVe} & {\bf NSLM} \\
\hline
\rule{0pt}{3ex}Numbers &  831.2 & 640.0 &  480.5 \\
\hspace{0.1in}\% change &  & (23.0\%) & (42.1\%) \\

\rule{0pt}{3ex}Locations & 4,969.4 & 3,888.3 & 1,833.6 \\
\hspace{0.1in}\% change &  & (21.8\%) & (63.1\%) \\
\hline

\end{tabular} & 
\end{tabular}
\caption{\textbf{Evaluation of Transfer from Larger Corpora.} Embeddings for Number and Geographic Locations word classes were initialized with GloVe embeddings trained on the 6B token Corpus with $d=300$.}
\label{tab:glove}
\end{table}

\subsection{Grounding with External Knowledge}

Limited amounts of external knowledge is provided to certain micro-models.  To evaluate the potential benefit of this extra knowledge, we experimented with \enquote{grounding} the word class embeddings \cite{Spithourakis2016NumericallyGL} by appending values for numbers and longitude/latitude coordinates for geographic locations, and setting these dimensions as non-trainable.  Models trained with this appended knowledge achieved non-trivial performance improvements over the NNLM baseline (see Table \ref{tab:grounded}), but are much smaller than the improvements due to micro-models.  To evaluate the combined effect of grounding and transfer from larger corpora we also evaluated our method using a model \textit{both} initialized with GloVe embeddings and grounded with external knowledge.  The NNLM component of these models improved number and geographic location perplexity by 26.1\% and 22.5\%, respectively, compared to a same-sized NNLM that was not provided with external knowledge or GloVe embeddings.  However, the NSLM component of these models performed better with perplexity improvements of 45.7\% and 63.3\%.

\begin{table}[h]
\centering
\small
\begin{tabular}{cccc}
\begin{tabular}{lrrr}
\hline
\rule{0pt}{3ex} {\bf } & {\bf NNLM}& {\bf Grounded}& {\bf \% change}\\\hline
\rule{0pt}{3ex}Numbers &  790.2 & 717.9 & (9.2\%)  \\
Global & 92.0 & 91.6 & (0.4\%) \\

\rule{0pt}{3ex}Locations & 4,812.1 & 3,890.2 & (19.2\%) \\
Global & 91.0 & 90.9 & (0.1\%)\\
\hline

\end{tabular} & 
\end{tabular}
\caption{\textbf{Summary Results for Grounded Embeddings.} The NNLM performs better with grounding, but the gains are smaller than with micro-models.}
\label{tab:grounded}
\end{table}

\iftrue
\subsection{Numbers and the Softmax Bottleneck}

Our results show that NNLMs have difficulty learning to predict numbers.  One possible explanation for this is the softmax bottleneck \cite{Yang2018BreakingTS}, which states that a standard softmax NNLM is rank limited by the number of hidden states of the model.  We would expect that correct numeric models are high-rank.  For example, encoding that a year observed in text should be numerically larger than the previous year entails a model with rank at least as large as the number of distinct years in the corpus.  However, we performed experiments on synthetic data that demonstrate that existing NNLMs can in fact approximate high-rank relationships on numbers almost perfectly, but the models perform relatively poorly because they fail to generalize.

\begin{table}[h]
\small
\begin{tabular}{ccccc}
\begin{tabular}{rrrrr}
\hline
\rule{0pt}{3ex}  & {\bf  } & {\bf NNLM}& {\bf GloVe} & {\bf NSLM} \\
{\bf $N$} & {\bf Training} & {\bf Test}& {\bf Test} & {\bf Test} \\
\hline
\rule{0pt}{3ex}100 &  1.0 & 99.1 & 37.5 & 1.0\\
200 &  1.0 & 203.5 & 60.0 & 1.0\\
1,000 &  1.0 & 1,021.0 & 463.1 & 1.0\\
10,000 &  1.0 & 10,030.4 & & 1.0\\
100,000 &  1.0 & 100,109.4 & & 1.0\\
\hline
\end{tabular} & 
\end{tabular}
\caption{\textbf{Evaluation of NNLM Capacity to Learn Increment Function.} Perplexities for Learning the Function $f(n)=n+1$.  NNLMs are able to {\em memorize} the input/output pairs, but are unable to generalize to the arithmetic function.}
\label{tab:bottleneck}
\end{table}

Specifically, we evaluate an LSTM of only 100 dimensions on the task of learning an increment function on a synthetic data set of consecutive pairs of integers represented as strings (see Table \ref{tab:bottleneck}).  The LSTM trains to 100\% accuracy even for tens of thousands of examples, demonstrating that the LSTM has the capacity to approximate a relationship with rank in the tens of thousands, meaning that the softmax bottleneck is not the limiting factor.   However, the LSTM is unsuprisingly unable to generalize to unseen pairs.  Even when using better number embeddings, from a GloVe model trained over a 6B token corpus, the model still generalizes poorly.  By contrast augmenting an NNLM with a simple micro-models allows the model to generalize well on this simple function.
\fi

\subsection{Experiments with AWD-LSTM}

We elected to evaluate NSLMs using a popular, standard architecture \cite{Zaremba2014RecurrentNN} that is fast to train.  To determine if our results were extensible, we also performed limited experiments on more expensive recent models.  For these experiments, we used AWD-LSTM models with 33 million parameters \cite{Merity2017RegularizingAO}.  There are three distinctions from the NSLMs presented above: a single model was used for both numbers and geographic word classes, models were trained for 500 epochs (with 750 being standard for this NNLM), and a single-task configuration was used meaning that instead of estimating $P_C^{(w)}$ for class labels, $P_C^{(w)}$ was calculated by summing all probability assigned to tokens in $V_C$.  These design choices were made for expediency.

\begin{table}[h]
\centering
\small
\begin{tabular}{cccc}
\begin{tabular}{lrrr}
\hline
\rule{0pt}{3ex} {\bf } & {\bf NNLM} & {\bf NSLM} & {\bf \% Change} \\
\hline
\rule{0pt}{3ex}Numbers & 421.1  & 336.4 & (20.1\%) \\
Locations & 2,066.3 & 1,340.7 & (35.1\%) \\
Global & 71.5  & 70.9 & (0.8\%) \\
\hline

\end{tabular} & 
\end{tabular}
\caption{\textbf{NSLMs with AWD-LSTM on Wikitext-2.}}
\label{tab:awdlstm}
\end{table}

Table Table \ref{tab:awdlstm} shows that NSLMs continue to offer significant perplexity improvements for this different architecture.  The improvements are consistent with Tables \ref{tab:numresults} and \ref{tab:georesults}, although somewhat smaller in percentage terms (which may be attributable to a ``floor effect'' as perplexity decreases).  We also note that global baseline perplexity is slightly higher than previously published results due to the re-tokenization of the corpora.

\section {Conclusion}

We introduced a novel language model architecture called NSLMs formed by composing a hierarchical NNLM with micro-models, and demonstrated that NSLMs substantially improve perplexity on number and geographic word classes.  We also introduced micro-models as a way to enhance the inductive bias of a language model by symbolically encoding domain knowledge.  We presented a procedure for constructing NSLMs for other domains, along with a discussion of how to formulate and select general-purpose building blocks for metric functions and probability density functions.  NSLMs are general in nature and can be formed from virtually any neural architecture with a softmax output layer.

Although we demonstrate the effectiveness of NSLMs on number and geographic word classes, our objective is not solely to achieve performance gains on a narrow set of word classes.   Language models inherently have an effective inductive bias which is well-suited to most running text.  We seek to establish a general method to tune language models for domain-specific tasks by enhancing the inductive bias with symbolic expressions that encode domain knowledge.

The domains explored in the paper lend themselves to numeric encoding and micro-models that perform simple operations on these encodings.  One topic of future research is to apply NSLMs to domains where symbolic expressions operate at a higher level of abstraction such as baseball play-by-plays or financial reporting.  Other enhancements to NSLMs may involve the use of the full preceding context encoded as the hidden state of a NNLM or using micro-models with multiple PDFs to allocate probability within a word class.

\section {Acknowledgements}

This work was supported in part by NSF Grant IIS-1351029.  The authors would like to thank Thanapon Noraset and the anonymous reviewers for their insightful comments and suggestions.

\bibliographystyle{aaai}
\bibliography{6053_aaai.bib}

\end{document}